\documentclass[10pt,twocolumn,letterpaper]{article}

\usepackage[pagenumbers]{iccv}

\definecolor{iccvblue}{rgb}{0.21,0.49,0.74}
\usepackage[pagebackref,breaklinks,colorlinks,allcolors=iccvblue]{hyperref}

\usepackage{xcolor}
\usepackage{lipsum}
\usepackage{tabularx}
\usepackage{multirow}
\usepackage{ulem}
\definecolor{lightgray}{rgb}{0.9,0.9,0.9}

\newcommand\nonumfootnote[1]{
\begingroup
    \renewcommand\thefootnote{}\footnote{\hspace{-3.5pt}#1}
    \addtocounter{footnote}{-1}
\endgroup
}

\newcommand{\projectpage}{\href{https://guoyww.github.io/projects/long-context-video/}{\textcolor{magenta}{Project Page}}\xspace}

\def\paperID{*****} 
\def\confName{ICCV}
\def\confYear{2025}

\title{Long Context Tuning for Video Generation}

\author{
Yuwei Guo$^{1,2}$ \quad
Ceyuan Yang$^{2,\dag}$ \quad
Ziyan Yang$^{2}$ \quad
Zhibei Ma$^{2}$ \quad
Zhijie Lin$^{2}$ \\ 
Zhenheng Yang$^{3}$ \quad
Dahua Lin$^{1}$ \quad
Lu Jiang$^{2}$
\vspace{1mm} \\
$^{1}$The Chinese University of Hong Kong \quad
$^{2}$ByteDance Seed \quad
$^{3}$ByteDance
}

\begin{document}

\maketitle

\nonumfootnote{$^\dag$Corresponding Author.}

\begin{abstract}

Recent advances in video generation can produce realistic, minute-long single-shot videos with scalable diffusion transformers.
However, real-world narrative videos require multi-shot scenes with visual and dynamic consistency across shots.
In this work, we introduce Long Context Tuning (LCT), a training paradigm that expands the context window of pre-trained single-shot video diffusion models to learn scene-level consistency directly from data.
Our method expands full attention mechanisms from individual shots to encompass all shots within a scene, incorporating interleaved 3D position embedding and an asynchronous noise strategy, enabling both joint and auto-regressive shot generation without additional parameters.
Models with bidirectional attention after LCT can further be fine-tuned with context-causal attention, facilitating auto-regressive generation with efficient KV-cache.
Experiments demonstrate single-shot models after LCT can produce coherent multi-shot scenes and exhibit emerging capabilities, including compositional generation and interactive shot extension, paving the way for more practical visual content creation.
See our \projectpage for more details.

\end{abstract}    
\begin{figure*}
    \centering
    \includegraphics[width=\linewidth]{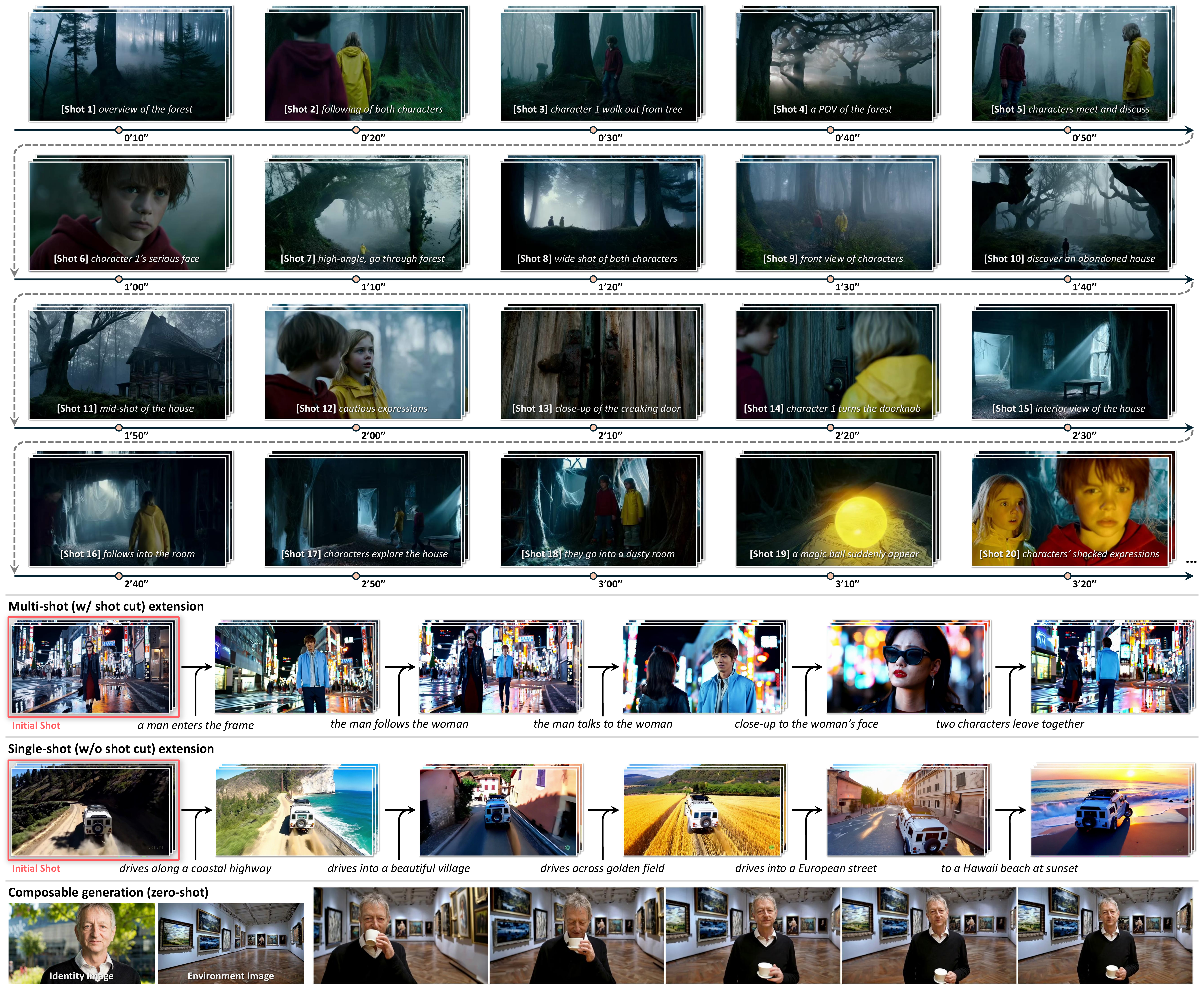}
    \caption{
    We propose Long Context Tuning~(LCT) to expand the context window of pre-trained single-shot video diffusion models.
    A direct application of LCT is scene-level video generation for short film production, as shown in the top example.
    We also show several emerging capabilities offered by LCT, such as interactive multi-shot direction and single shot extension, as well as zero-shot compositional generation, despite the model having never been trained on such tasks.
    \textit{We recommend referring to our \projectpage for better visualization.}
    }
    \label{fig:teaser}
\end{figure*}

\section{Introduction}

Video generation has experienced significant advancements in recent years.
By leveraging web-scale data and scalable model architecture such as diffusion transformer~(DiT)~\cite{peebles2023scalable}, state-of-the-art models~(\eg, SoRA~\cite{videoworldsimulators2024}, Kling~\cite{kling}, Gen3~\cite{gen3}) are now capable of synthesizing realistic single-shot videos lasting up to a minute.
However, real-world narrative videos, such as movies and television shows, are composed of multiple single-shot segments.
This implies a substantial gap between the current capabilities of single-shot video generation and the authentic demands of video content production.

To bridge this gap, video generation research may need to evolve from single-shot synthesis to scene-level generation.
In this higher-level paradigm, a scene is defined as a series of single-shot videos capturing coherent events unfolding over time~\cite{rasheed2003scene, chasanis2008scene, rao2020local}.
For instance, the classic scene in \textit{Titanic} where \textit{Jack} and \textit{Rose} meet on the deck consists of four principal shots~(\cref{fig:data_example}):
(1)~close-shot of \textit{Jack} looking back,
(2)~mid-shot of \textit{Rose} speaking,
(3) wide-shot of \textit{Rose} approaching \textit{Jack}, and
(4) close-up shot of \textit{Jack} embracing \textit{Rose} from behind.
Generating such a scene requires consistency in \textit{visual appearance} and \textit{temporal dynamics} to ensure a coherent narration flow.
Specifically, visual coherence indicates the consistent rendering of shared visual components, \eg, person identity, background, lighting, and color tone.
Similarly, temporal coherence necessitates the uniformity in dynamic elements such as character actions
(\eg, maintaining consistent walking pace) 
and camera movements (\eg, smooth versus shaky camera) across shots.
Addressing these challenges demands novel approaches for scene-level video generation.

Many solutions have been proposed to address scene-level video generation, most of which fall into two categories:
(1)~appearance-conditioned generation~\cite{long2024videostudio, huang2025conceptmaster, jiang2024videobooth}, and (2)~keyframe generation~\cite{zhao2024moviedreamer, zheng2024videogen, zhou2024storydiffusion, huang2024context} followed by image-to-video~(I2V) animation~\cite{xing2024dynamicrafter, blattmann2023stable}.
In the first paradigm, key visual elements (\eg, character identity and background) serve as conditional inputs to enforce cross-shot consistency.
However, this approach struggles to maintain abstract elements like lighting and color tone due to its reliance on predefined conditions and specially curated datasets.
The keyframe-based strategy generates a coherent set of keyframes to ensure visual consistency, which then serve as initial frames for an I2V model to synthesize each shot independently.
Yet, the independent shot synthesis fails to guarantee temporal consistency across shots.
Moreover, sparse keyframes limit the effectiveness of conditioning.
For instance, if a character enters the scene between keyframes, the keyframe-based method misses this character as part of the conditioning, thereby rendering a consistent representation of the character is infeasiable.

In this work, we propose \emph{Long Context Tuning (LCT)} for scene-level video generation.
LCT builds upon a pre-trained single-shot video generation model, adapting it to learn a longer context window capable of modeling cross-shot consistency from scene-level video data.
To achieve this, we introduce three novel design elements. First, LCT adapts the full attention mechanism from individual shots to encompass all shots within a scene.
Inspired by \cite{wang2024qwen2}, we propose an interleaved 3D Rotary Positional Embedding (RoPE)~\cite{su2024roformer} that assigns unique absolute positions to individual shots while maintaining the relative positional relationships among text-video tokens within each shot.
Second, we unify visual condition inputs and diffusion samples by applying independent diffusion timesteps to each shot.
This strategy enables joint denoising of all shots, or using some as conditions by setting their noise to a low level.
Finally, we demonstrate the model with bidirectional attention after LCT can be further fine-tuned to context-causal attention, which facilitates efficient auto-regressive generation using KV-cache and substantially reduces computational overhead.

After training on the scene-level video data, experimental results demonstrate that our model exhibits outstanding performance in generating visually and semantically consistent scenes. As shown in Fig.~\ref{fig:teaser}, we highlight a generated video with around 20 shots that lasts 3 minutes and keep the appealing visual and semantic consistency.
Notably, we find the LCT exhibits emergent capabilities that go beyond those offered by the pre-trained model it builds upon.
For instance, when provided with character identity and environment image (\emph{e.g.,} the bottom of~\cref{fig:teaser}), the model can perform compositional generation by synthesizing videos that seamlessly integrate these elements, despite not being explicitly trained for this task.
Moreover, the LCT model also supports autoregressive shot extension, both with and without shortcuts (\emph{e.g.,} the woman walking and car driving examples in~\cref{fig:teaser} respectively).
This feature is particularly useful because it divides long video generations into scene clips, which enables interactive modification by human users.
We anticipate that this work could bring inspiration for future research in long video generation.

\section{Related Works}

\noindent\textbf{Shot-level Video Generation.}
Previous literature has extensively explored single-shot video generation.
Early approaches~\cite{tulyakov2018mocogan, skorokhodov2022stylegan, wang2020imaginator, chu2020learning} relied on GANs but were limited to single-domain datasets.
The field later shifted to diffusion-based methods~\cite{ho2022video, zhang2024show, ge2023preserve, chen2024videocrafter2, chen2023videocrafter1, wang2024lavie, luo2023videofusion, zhou2022magicvideo, zeng2024make}, mostly leveraging pre-trained image diffusion models.
Among them, representative works such as Align-Your-Latents~\cite{blattmann2023align}, AnimateDiff~\cite{guo2023animatediff}, Stable Video Diffusion~\cite{blattmann2023stable}, Lumiere~\cite{bar2024lumiere}, and Emu Video~\cite{girdhar2024factorizing} extend 2D diffusion UNet with temporal layers to model dynamic motion priors.
Later, SoRA~\cite{videoworldsimulators2024} demonstrated the scalability of diffusion transformers (DiT)~\cite{peebles2023scalable} by significantly advancing video generation quality through 3D autoencoding and full-attention mechanisms~\cite{vaswani2017attention}.
This inspired subsequent developments including commercial models like Kling~\cite{kling}, Gen3~\cite{gen3}, and Seaweed, as well as open-source alternatives CogVideoX~\cite{yang2024cogvideox}, Mochi~\cite{genmo2024mochi}, and HunyuanVideo~\cite{kong2024hunyuanvideo}.
Parallel approaches explore streaming generation~\cite{yin2024slow, henschel2024streamingt2v} and decoder-only language models~\cite{kondratyuk2023videopoet} using discretized visual tokens~\cite{yu2023magvit, yu2023language}.
Our approach is compatible with single-shot video diffusion models with DiT-based architecture.

\vspace{0.5em}
\noindent\textbf{Scene-level Video Generation.}
Recent works have addressed scene-level video generation, which requires synthesizing sequential videos depicting continuous events~\cite{hu2024storyagent, xie2024dreamfactory, bansal2024talc, yang2024synchronized}.
Research in this field primarily falls into two categories: appearance-conditioned approaches~\cite{huang2025conceptmaster, jiang2024videobooth} and keyframe-based methods.
The former enforces consistency by conditioning on character identity or environment images, while the latter jointly generates initial keyframes before applying independent image-to-video (I2V)~\cite{xing2024dynamicrafter, zhang2023i2vgen}.
VideoStudio~\cite{long2024videostudio} extends text prompts with entity embeddings to preserve appearance information, while MovieDreamer~\cite{zhao2024moviedreamer} predicts coherent visual tokens that are rendered into I2V keyframes via diffusion decoders. 
VGoT~\cite{zheng2024videogen} structures multi-shot generation using identity-preserving embeddings to maintain cross-shot consistency.
Other approaches explore feature sharing~\cite{atzmon2024multi} or retrieval augmentation~\cite{he2023animate, wang2024dreamrunner}.
Nevertheless, existing methods still struggle to model complex scene-level coherence, and keyframe-based approaches are prone to suffer from the ineffective conditioning.

\vspace{0.5em}
\noindent\textbf{Long Video Generation.}
Current single-shot video models are limited to generating ten-second clips from single prompts.
Research addressing this constraint by exploring training-free methods~\cite{lu2025freelong, wang2023gen}, auto-regressive frameworks~\cite{yin2024slow}, hierarchical strategy~\cite{he2022latent, yin2023nuwa}, as well as scale up via distributed inference~\cite{tan2024video}.
Among them, FreeNoise~\cite{qiu2023freenoise} reschedules the noise sequence and performs window-based temporal attention.
StreamingT2V~\cite{henschel2024streamingt2v} and CasusVid~\cite{yin2024slow} extend the video length via auto-regressive frame generation.
Recent advances incorporate additional control mechanisms~\cite{villegas2022phenaki, oh2024mevg, cai2024ditctrl}, including MinT's~\cite{wu2024mind} time-dependent prompts with specialized position embeddings, as well as DFoT~\cite{song2025history, chen2025diffusion}'s history-guided video frame rolling out.
Our approach, by contrast, directly enables long video extension via shot extension with and without shotcut.

\section{Method}

Our goal is scene-level video generation that synthesizes multi-shot videos with consistency.
To this end, we propose \textit{Long Context Tuning~(LCT)} upon the pre-trained single-shot video diffusion model to expand its context window and learn scene-level correlation directly from data.

In~\cref{sec:single_shot_model}, we provide preliminaries on single-shot video models.
In~\cref{sec:towards_long_context}, we detail our approach for LCT.
In~\cref{sec:causal_finetuning}, we explore context-causal attention fine-tuning for efficient auto-regressive generation.
In~\cref{sec:implementation}, we discuss training and inference implementation.

\subsection{Preliminary: Single-shot Video Models}
\label{sec:single_shot_model}

We build our method upon a latent video diffusion transformer~(DiT)~\cite{peebles2023scalable} model.
The diffusion process operates on the latent representation by encoding RGB video $x_0$ into $z_0 = \mathcal{E}(x_0) \in \mathbb{R}^{c'\times h'\times w'\times f'}$.
The model is trained with Rectified Flow~(RF) formulation~\cite{lipman2022flow, liu2022flow, esser2024scaling}, where the noisy sample is a linear interpolation between clean data point and sampled Gaussian noise $\epsilon$, \ie, $z_t = (1 - t)z_0 + t\epsilon$.
The training objective is to regress the velocity field, \ie,
\begin{equation}\label{eq:loss}
    \mathcal{L} = \mathbb{E}_{t, z_0, \epsilon}\lVert v_\Theta(z_t, t, c_{text}) - (\epsilon - z_0) \rVert_2^2,
\end{equation}
where $t \in [0, 1]$ is the continuous diffusion timestep, $v_\Theta(\cdot)$ is the neural network, $c_{text}$ is the text prompt condition.

The single-shot model employs an MMDiT~\cite{esser2024scaling} design, whose transformer block has seperate sets of weights for text and video tokens but joins the sequence of the two modalities for self-attention operation.
To encode position information, 3D Rotary Position Embedding~(RoPE)~\cite{su2024roformer} is applied to the video tokens, where the axis of height, width, and frame index are encoded to different latent channels.

\subsection{Towards Long Context Video Generation}
\label{sec:towards_long_context}
We detail the core components of LCT for scene-level video generation in this section, including data preparation~(\cref{sec:data_curation}) and architecture design~(\cref{sec:learn_beyond_single_shot}).
We further unify conditioning inputs and diffusion samples via an asynchronous training timesteps strategy~(\cref{sec:unify_condition}).

\begin{figure}
    \centering
    \includegraphics[width=\linewidth]{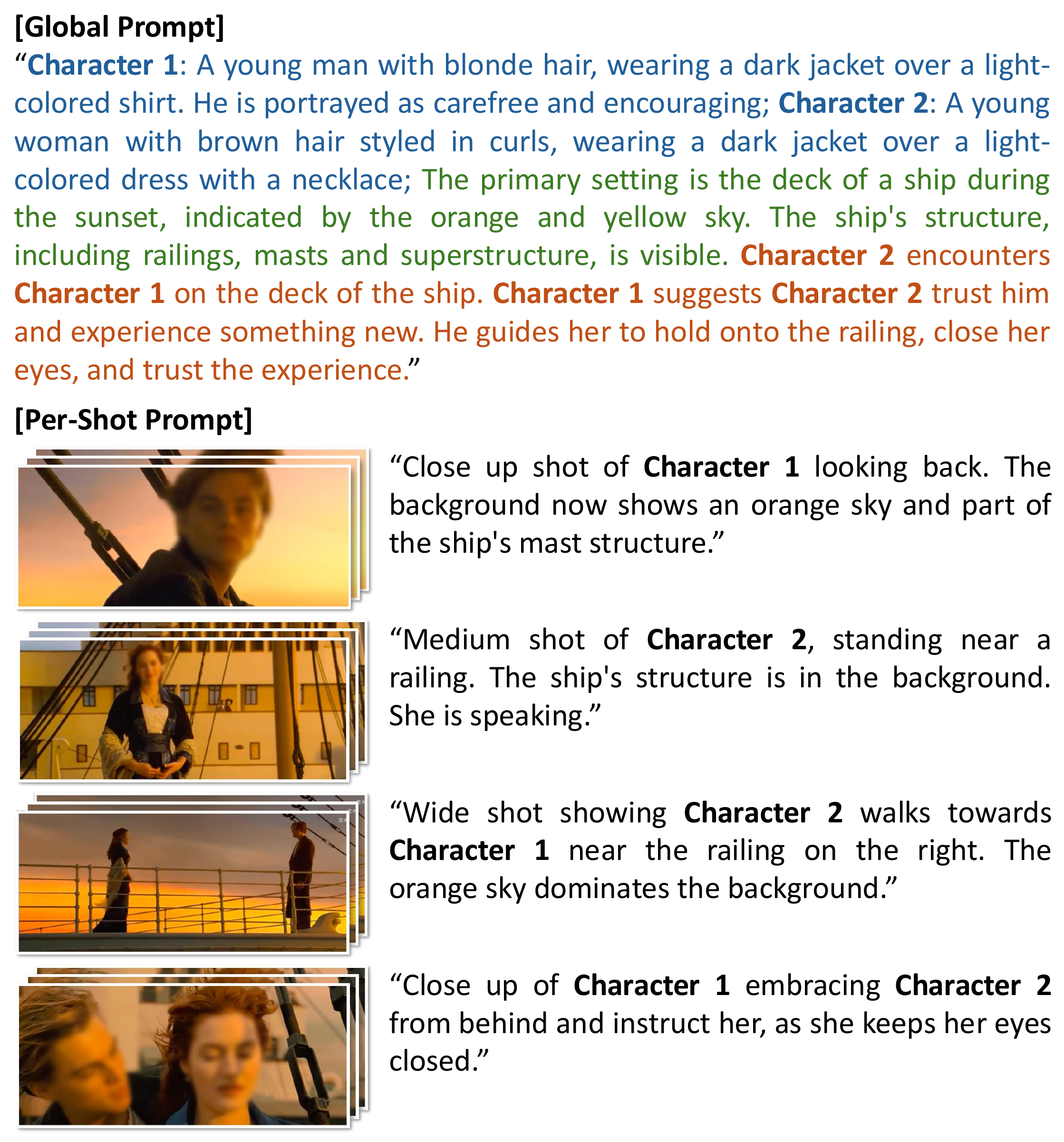}
    \caption{
    \textbf{Scene-level Video Data Example.}
    \textit{Global prompt} contains shared elements like \textcolor[RGB]{33,95,154}{\bf character}, \textcolor[RGB]{59,125,35}{\bf environment}, and \textcolor[RGB]{192,80,21}{\bf story overview}, while
    \textit{per-shot prompt} details events in each shot.
    }
    \label{fig:data_example}
\end{figure}

\subsubsection{Data Curation}
\label{sec:data_curation}

\noindent\textbf{Prompt Structure Definition.}
This paper considers scene as a coherent sequence of video clips sharing high-level semantic concepts (characters, environment) while varying in shot-level features (view angles, temporal events).
Therefore, we employ a two-tier prompt structure: \textit{global prompts} and \textit{per-shot prompts}, as shown in~\cref{fig:data_example}.
Specifically, \textit{global prompts} capture shared elements across shots, following the format ``\texttt{[Character] [Environment] [Story]}", where character descriptions use the structure of ``\texttt{Character [ID]: [Description]}".
\textit{Per-shot prompts} detail specific events within each shot and reference characters using ``\texttt{Character [ID]}", rather than common but ambiguous descriptors like ``the man/woman".

\vspace{0.5em}
\noindent\textbf{Data Processing.}
We collect scene-level video data from public sources across various genres (movies, documentaries, \etc).
Raw videos are first segmented into scene videos using scene boundary detection algorithms, then further divided into individual shots via shot-cut detection.
We feed each scene video to Gemini-1.5~\cite{team2024gemini} and prompt it to provide both global and shot-level descriptions in our specified format.
This process yields approximately 500K scene samples averaging 5 shots each.
To supplement the training data, we filter some single-shot videos with large temporal variants, divide them into sub segments based on the event changes, and treat them as multi-shot videos.
Therefore, adjacent shots transition smoothly without abrupt shotcut, and this process contributes roughly 1M additional samples.
We add ``\texttt{[SHOT CUT]}" before shot-level prompt in the authentic multi-shot dataset as a special mark.

\begin{figure*}
    \centering
    \includegraphics[width=\linewidth]{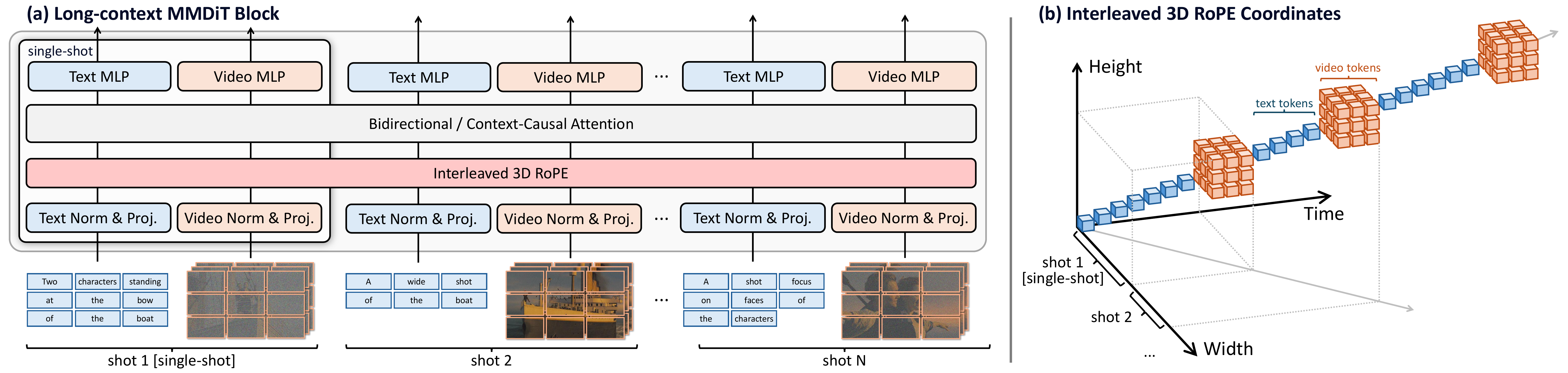}
    \caption{
    \textbf{Architecture Designs.}
    (a)~\textit{Long-context MMDiT block}.
    We expand the attention operation to all text and video tokens within a scene, and apply independent noise levels to individual shots.
    The interleaved 3D RoPE assigns distinct coordinates for each shot.
    (b) \textit{Interleaved 3D RoPE coordinates.}
    At shot-level, text tokens precede video tokens along the space diagonal.
    At scene-level, tokens are arranged shot by shot, forming an interleaved ``\texttt{[text]-[video]-[text]-...}" pattern along the space diagonal.
    }
    \label{fig:method}
\end{figure*}

\subsubsection{Learning Beyond Single-shot}
\label{sec:learn_beyond_single_shot}

\noindent\textbf{Long-term Modeling via Full Attention.}
Avoiding introducing additional inductive bias, we choose the vanilla full attention~\cite{peebles2023scalable, vaswani2017attention, videoworldsimulators2024} to model scene-level consistency.
This is done by joining all text and video tokens within the scene, and performing attention operation on the combined sequence jointly, as shown in the Long-context MMDiT in~\cref{fig:method}~(a).
Note that when the scene only contains one video, the Long-context MMDiT degrades to a single-shot MMDiT.
Thus, this framework is still compatible with single shot generation and can be trained with single-shot data to preserve the pre-trained capability.

\vspace{0.5em}
\noindent\textbf{Interleaved Position Embedding.}
Directly joining all tokens within a scene will hinder the model from identifying which shot the tokens belong to, since tokens from different shots are treated equally.
We resolve this issue via an interleaved 3D position embedding, which is conceptually similar to M-RoPE~\cite{wang2024qwen2} but has not been tested in the diffusion models literature.
Specifically, we first add 1D-equivalent 3D RoPE~\cite{su2024roformer} for text-tokens by setting corrdinates on three axis identical, and place the video tokens after the end of the text tokens along the space diagonal, as shown in the ``shot~1 (single shot)" part in~\cref{fig:method}~(b).
This is done via efficient single-shot model fine-tuning.

When deal with multiple shots, we keep single-shot's relative text-video tokens position and append the text-video token group shot by shot, forming an interleaved ``\texttt{[text]-[video]-[text]-...}" token sequence along the space diagonal, as shown in~\cref{fig:method}~(b).
Keeping the relative text-video position allows each shot to inherit the text-visual alignment from the pre-trained model, while different absolute positions distinguish the relation between tokens and the corresponding shot.
To deal with global prompt in~\cref{sec:data_curation}, we add dummy video tokens and treat it as a normal text-video pair.

\subsubsection{Unify Conditions and Diffusion Samples}
\label{sec:unify_condition}

While the model can generate scene-level multi-shot videos from text prompts with the proposed designs, incorporating visual conditioning enables valuable capabilities like story extension and auto-regressive generation.
Unlike existing approaches that rely on auxiliary networks for visual conditioning~\cite{zhang2023adding, ye2023ip, guo2024sparsectrl, he2024cameractrl}, we unify conditioning inputs and diffusion samples through an asynchronous timestep strategy.
As shown in~\cref{fig:method}~(a), we assign independently sampled diffusion timesteps to each shot during training, rather than applying uniform timesteps across all shots.
This establishes dynamic dependencies between shots and encourages the model to exploit cross-shot relationships more effectively.
For instance, when a shot within the context window exhibits lower noise compared to others, it naturally serves as a rich source of appearance information to guide the denoising process of noisier shots.
Consequently, we can set some samples' noise to a low level to utilize them as visual conditions, or synchronize diffusion timesteps across all samples for joint generation~(\cref{fig:inference}~(a)).

\subsection{Causal Attention Fine-tuning}
\label{sec:causal_finetuning}

\begin{figure}[t]
    \centering
    \includegraphics[width=\linewidth]{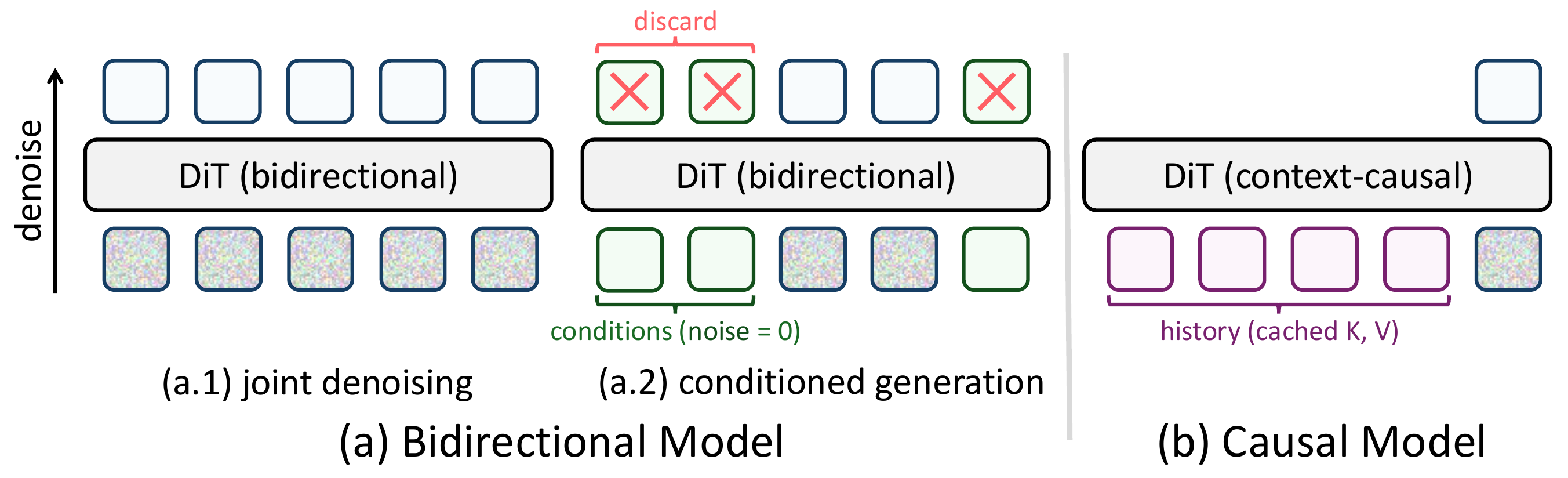}
    \vspace{-1.5\baselineskip}
    \caption{
    \textbf{Inference Modes.}
    (a)~\textit{Bidirectional} model enables (a.1)~joint or (a.2)~visual-conditioned generation, while (b)~\textit{context-causal} model supports auto-regressive generation.
    }
    \label{fig:inference}
\end{figure}

Our conditioning mechanism enables auto-regressive shot generation by using cleaner history samples as conditions.
In this paradigm, information flow is inherently \textit{directional}, \ie, cleaner history samples require little information from subsequent noisy samples, while noisy samples extract cues from preceding history to ensure consistency.
This suggests that \textit{bidirectional} attention is redundant and can be replaced with more efficient \textit{causal} attention.
After training the bidirectional model with LCT, we therefore implement context-causal attention (bidirectional within each shot, but with tokens attending only to their preceding context) and fine-tune this causal variant.
During inference, this architecture allows K, V features cached from history sample generation, eliminating repeated computation and thereby significantly reducing computational overhead~(\cref{fig:inference}~(b)).

\subsection{Implementations}
\label{sec:implementation}

\noindent\textbf{Training.}
We jointly train the model on single-shot and scene-level video data to preserve its pre-trained capability.
To enable image generation/conditioning, we randomly substitute shots with one of its single frame at a predetermined probability.
Diffusion timesteps are independently sampled from a logit-normal distribution for each shot, with per-shot losses computed according to~\cref{eq:loss} and averaged before gradient backpropagation.

\vspace{0.5em}
\noindent\textbf{Multi-shot Generation with Human Selection.}
Cross-shot dependencies in a scene are often non-sequential, \ie, a shot may refer to elements from distant earlier shots rather than just recent ones.
For instance, a character reappearing after several shots requires conditioning on their initial appearance rather than merely on adjacent shots, departing from common auto-regressive generation that is vulnerable to errors accumulation.
This insight enables our generation strategy with human selection.
We first establish a history pool by generating key character and environmental context shots, then selectively draw from this pool when generating new shots based on relevance rather than recency.

\section{Experiments}

\begin{figure*}
    \centering
    \includegraphics[width=\linewidth]{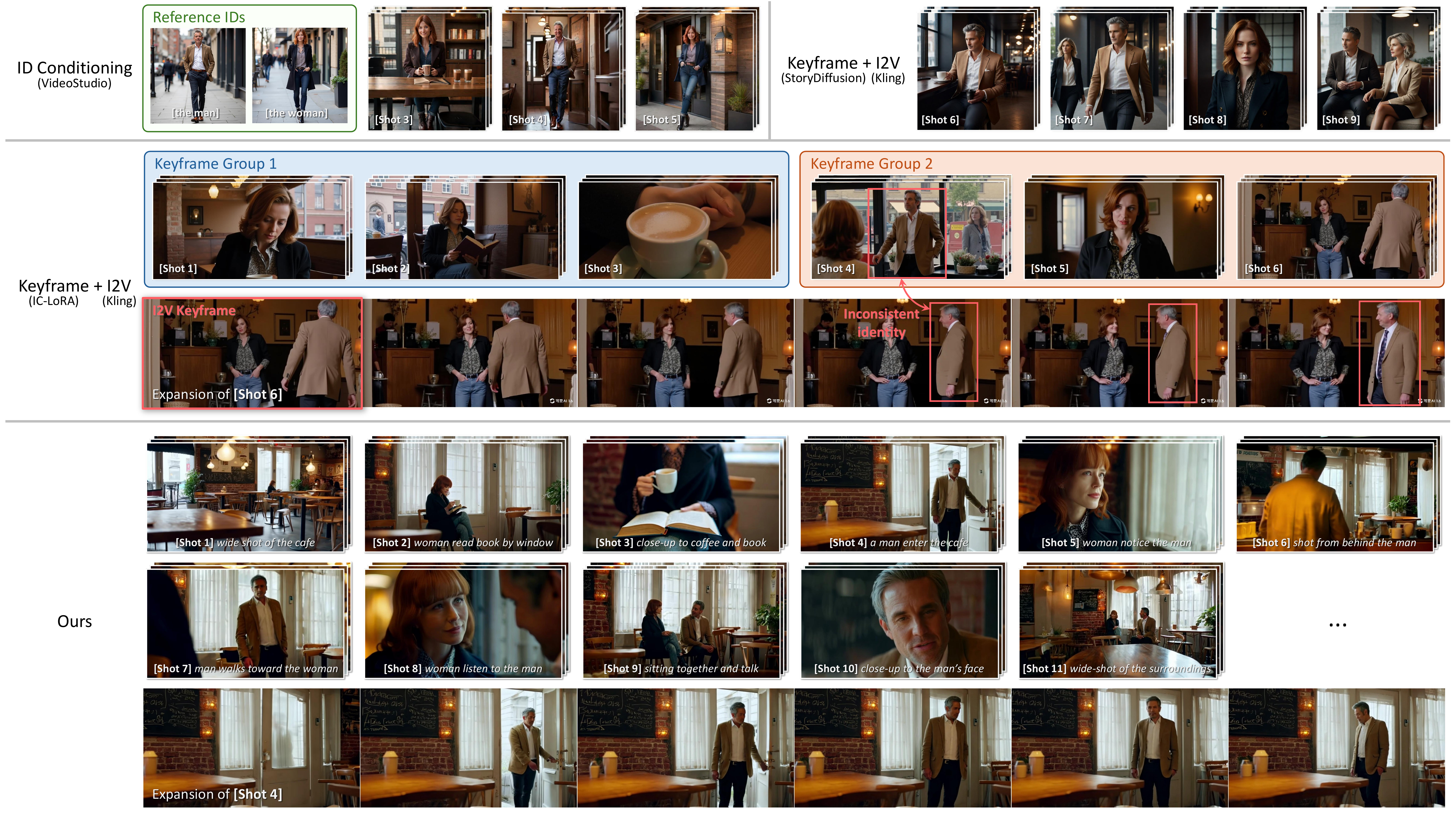}
    \caption{
    \textbf{Qualitative Comparisons.}
    We show stacked video frames synthesized by all methods, and expand two shots to illustrate the ``reappearance" issue discussed in~\cref{sec:comparison}.
    The simplified prompts for each shot can be found in the subtitle in ``Ours".
    }
    \label{fig:qualitative_compare}
\end{figure*}

\vspace{0.5em}
\noindent\textbf{Training Details.}
We implement LCT on a text-to-video diffusion model that adopts MMDiT~\cite{esser2024scaling} architecture and is trained on images and videos in their native resolutions and durations~\cite{dehghani2023patch}.
The parameters scale of the pre-trained model is 3B.
We use a context window size of nine shots in maximum, and continue the model training with LCT on 128 NVIDIA H800s for 135K iterations.
For causal attention fine-tuning, we load the model weights after LCT and fine-tune for 9K iterations.
The training resolution equals the area size of $480 \times 480$ with untouched aspect ratio.

\subsection{Comparison}
\label{sec:comparison}

\noindent\textbf{Settings.}
We compare our approach against previous arts on scene-level video generation.
We focus on two major categories of existing solutions: (1)~\textit{appearance-conditioned} approaches, where we adopt VideoStudio~\cite{long2024videostudio};
and (2)~\textit{keyframe-based} apporaches, where we employs two keyframe generation approaches, \ie, FLUX~\cite{flux} storyboard In-Context LoRA~(IC-LoRA)~\cite{huang2024context} and StoryDiffusion~\cite{zhou2024storydiffusion}, as well as Kling~\cite{kling} for image-to-video~(I2V) animation.
Since IC-LoRA only synthesizes three images, we divide a scene into multiple groups and generate keyframes separately.
We prompt o3-mini~\cite{o3mini} to design the scene and transform the text prompt into method-specific format.

\vspace{0.5em}
\noindent\textbf{Qualitative Results.}
The qualitative results in~\cref{fig:qualitative_compare} reveal several notable observations.
First, baseline methods exhibit limited frame composition diversity.
In VideoStudio, character poses closely mirror reference images, while IC-LoRA and StoryDiffusion predominantly generate character-centric mid-shots.
Our approach, however, produces diverse framing across wide-shots~(\eg, shots 1, 11), mid-shots~(\eg, shots 2, 4, 6), and close-up shots~(\eg, shots 3, 8, 10), enabling richer narrative perspectives.
Second, baselines demonstrate deficient prompt alignment.
IC-LoRA notably fails to generate an establishing shot~(\ie, environment without characters) for shot 1, instead introducing extraneous characters in other shots.
This stems from concatenating shot descriptions into a single prompt, impeding the model to disentangle semantic elements.
VideoStudio and StoryDiffusion tend to synthesize characters' front-facing views and thus compromise prompt alignment.
Third, our method achieves superior consistency.
The female character maintains her position beside the window throughout the scene, whereas baselines show disruptive positional shifts that break narrative continuity.

\begin{table}[t]
    \centering
    \resizebox{\linewidth}{!}{
    \begin{tabular}{cccccc}
    \toprule
    \multirow{2}{*}{Method} & \multicolumn{2}{c}{\textit{Visual}} & \textit{Temporal} & \multicolumn{2}{c}{\textit{Semantic}} \\
    & Aesthetic$_\uparrow$ & Quality$_\uparrow$ & Consistency~(avg.)$_\uparrow$ & Text$_\uparrow$ & User Study$_\uparrow$ \\
    \midrule
    VideoStudio~\cite{long2024videostudio}& \textbf{61.68} & \uline{73.13} & 95.25  & \uline{28.00} & 2.14 \\
    StoryDiffusion~\cite{zhou2024storydiffusion} + Kling~\cite{kling} & 60.40 & \textbf{74.04} & \textbf{96.57} & 27.33 & \uline{2.50} \\
    IC-LoRA~\cite{huang2024context} + Kling~\cite{kling} & 57.88 & 69.07 & \uline{96.27} & 27.90 & 1.57 \\
    Ours & \uline{60.79} & 67.44 & 95.65 & \textbf{30.14} & \textbf{3.79} \\
    \bottomrule
    \end{tabular}
    }
    \caption{\textbf{Quantitative Evaluations.}
    We adopt automatic metrics and average human ranking~(AHR).
    ``Consistency~(avg.)" represents the average score of subject and background consistency.
    }
    \label{tab:quantitative_result}
\end{table}

Finally, we highlight an often overlooked ``reappearance" issue inherent in keyframe-based scene-level video generation approaches.
As evident in the ``Expansion of [Shot 6]" in the ``Keyframe~(IC-LoRA) + I2V~(Kling)" approach, when a character's identity is not fully captured in the initial keyframe but appears in subsequent I2V video frames, its consistency across shots cannot be guaranteed.
Conversely, our direct video generation solution effectively eliminates this problem, as demonstrated in the ``Expansion of [Shot 4]" in our results.

\vspace{0.5em}
\noindent\textbf{Quantitative Results.}
We evaluate all methods across three dimensions: visual quality, temporal consistency, and semantic coherence.
For automatic assessment, we utilize VBench~\cite{huang2024vbench} that focuses on shot-level quality.
Given the absence of scene-level coherence metrics, we employ user study and instruct participants to prioritize cross-shot consistency over appearance.
As shown in~\cref{tab:quantitative_result}, while our method performs slightly below visually-conditioned baselines in visual quality, it substantially outperforms them in semantic alignment, demonstrating superior capability in maintaining scene-level coherence.

\subsection{Ablative Studies}
\label{sec:ablation}

In this section, we discuss the model's training and inference behaviors of Long Context Tuning~(LCT).

\vspace{0.5em}
\noindent\textbf{Single-shot Generation after LCT.}
Our proposed LCT introducing \textit{no additional parameters}, enlarging the model capability under the same capacity.
Therefore, it is important to evaluate whether LCT sacrifices the pre-trained single-shot generation ability.
We investigate this by synthesizing videos using an established prompt suite with models before and after LCT, and evaluate them with VBench~\cite{huang2024vbench}.
As shown in the upper part of~\cref{tab:quantitative_ablation}, the post-LCT model demonstrates consistent performance gains in all metrics, indicating that LCT not only maintains but potentially enhances single-shot generation capability.

\begin{table}[t]
    \centering
    \resizebox{\linewidth}{!}{
    \begin{tabular}{ccccccc}
    \toprule
    Task & Model & Settings & Aesthetic$_\uparrow$ & Quality$_\uparrow$ & Consistency~(avg.)$_\uparrow$ & Text$_\uparrow$ \\
    \midrule
    \multirow{2}{*}{single-shot} & pre-trained & - & 52.09 & 63.37 & 94.98 & 30.81 \\
    & post-LCT & - & 55.91 & 64.23 & 95.06 & 32.22 \\
    \midrule
    \multirow{3}{*}{multi-shot} & bidirectional & joint denoise & 56.14 & 57.68 & 95.34 & 29.33 \\
    & bidirectional & auto-regressive & 55.46 & 56.62 & 94.87 & 30.24 \\
    & context-causal & auto-regressive & 55.49 & 50.31 & 94.96 & 30.15 \\
    \bottomrule
    \end{tabular}
    }
    \caption{\textbf{Quatitative Ablation Studies.} We compare single-shot generation capabilities in the upper part, and the effects of inference modes in the lower part.}
    \label{tab:quantitative_ablation}
\end{table}

\vspace{0.5em}
\noindent\textbf{Effects of Inference Modes.}
Our approach supports multiple inference modes: joint generation and auto-regressive~(conditional) generation for bidirectional model~(\cref{fig:inference}~(a)), as well as auto-regressive generation for the context-causal model~(\cref{fig:inference}~(b)).
To assess how they affect generation quality, we evaluate videos produced by each mode.
The lower section of~\cref{tab:quantitative_ablation} reveals the improved text alignment from joint generation to auto-regressive settings.
This likely occurs because in auto-regressive modes, later shots can access preceding history as complementary information to the text prompts.

\begin{figure}[t]
    \centering
    \includegraphics[width=\linewidth]{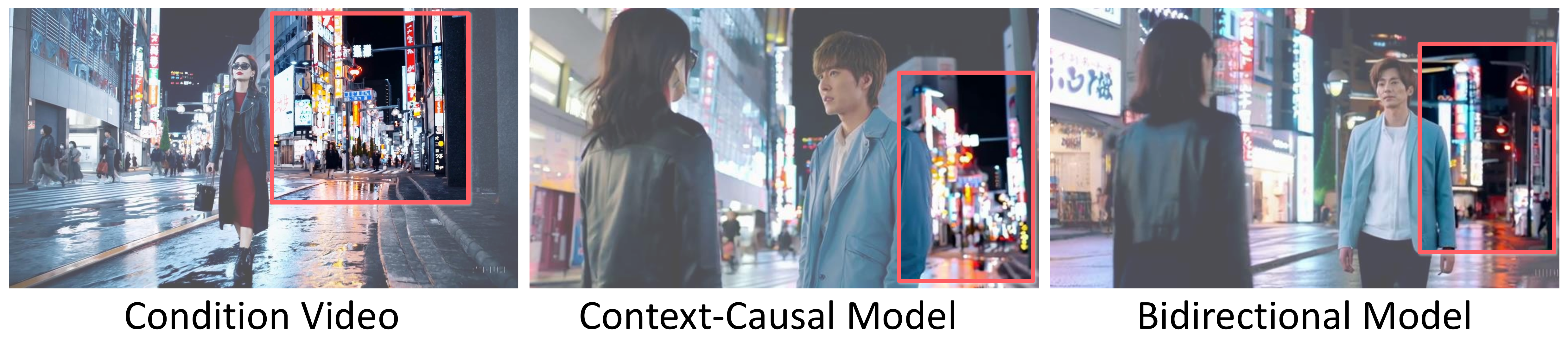}
    \caption{\textbf{Fidelity to History Condition}.
    The video background generated by the causal model exhibits superior fidelity to the history condition, as evidenced by the street lights' layout.
    }
    \label{fig:ar_fidelity}
\end{figure}

Moreover, we found that the context-causal architecture demonstrates superior fidelity to history conditions compared to the bidirectional one, as evident in~\cref{fig:ar_fidelity}.
We attribute this to the causal attention mechanism, which enforces sequential dependency on history and consequently assigns greater weight to preceding conditions.

\begin{figure}
    \centering
    \includegraphics[width=\linewidth]{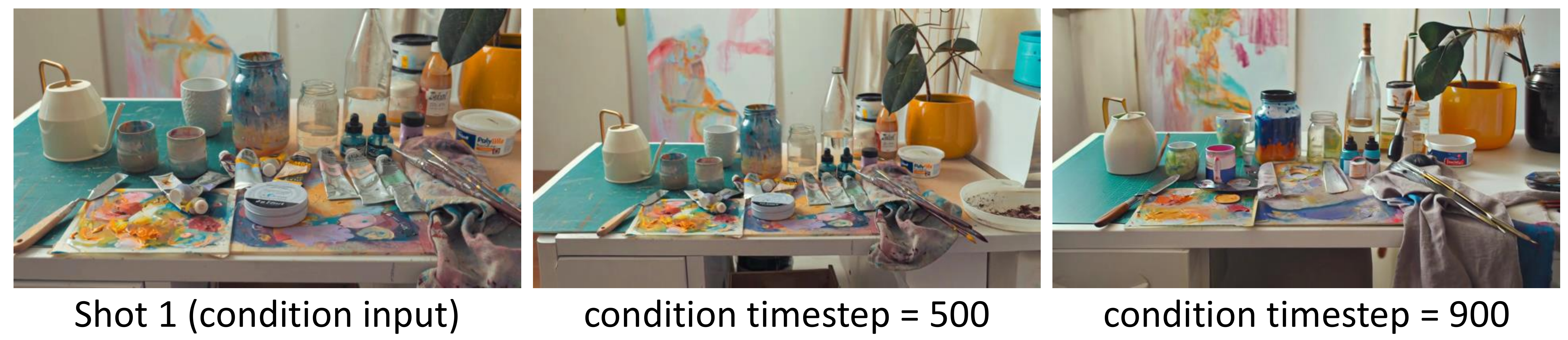}
    \caption{\textbf{Effects of Conditioning Timestep.} Large timesteps sacrifice fidelity to the condition.}
    \label{fig:condition_noise}
\end{figure}

\begin{figure}
    \centering
    \includegraphics[width=\linewidth]{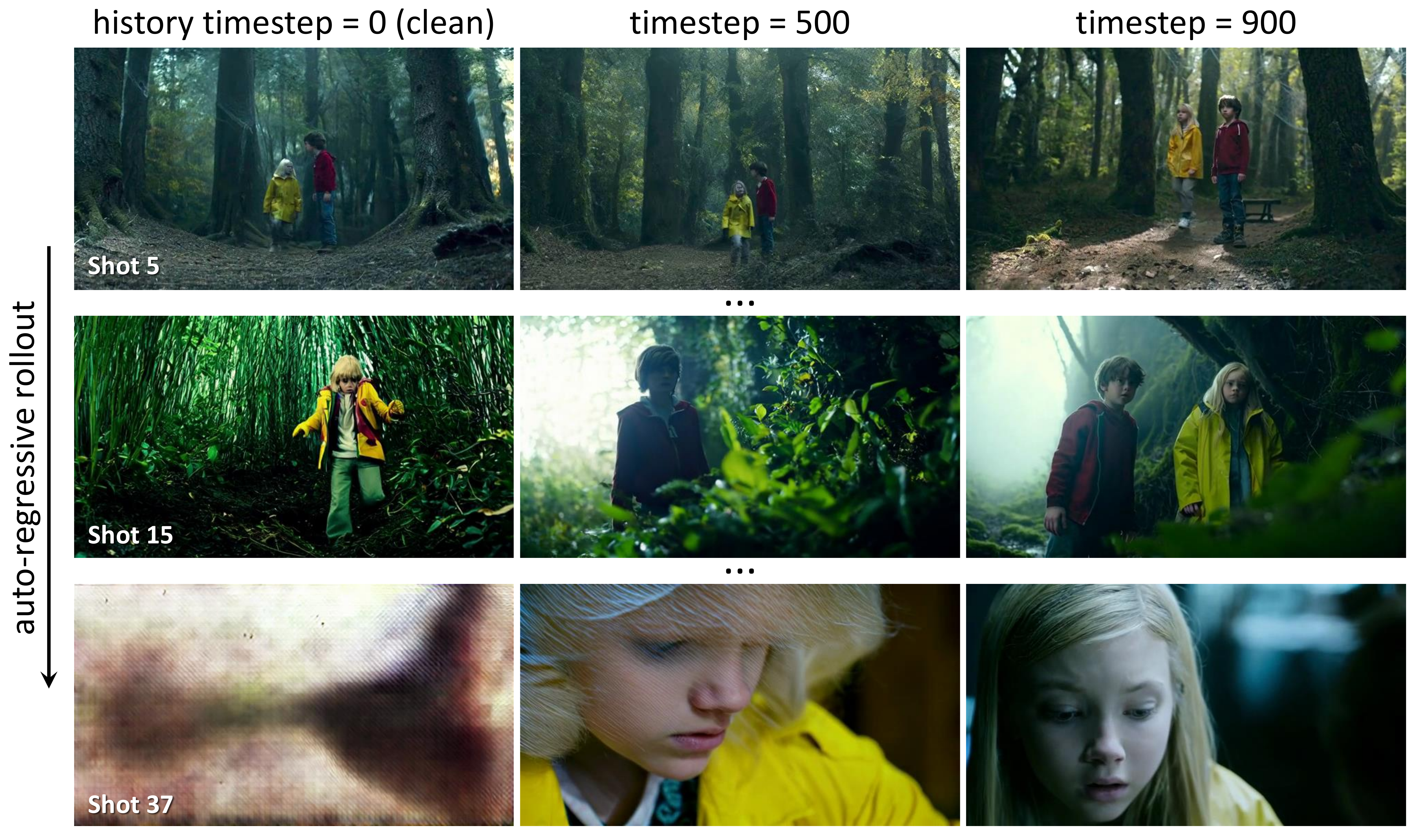}
    \caption{\textbf{Effects of History Timestep.} Large timesteps mitigate ``error accumulation" issue at the cost of history fidelity.}
    \label{fig:ar_noise}
\end{figure}

\vspace{0.5em}
\noindent\textbf{History Conditioning Timestep.}
In auto-regressive inference, previously generated samples are perturbed to noise level $t_c$ and utilized as conditioning inputs, as shown in~\cref{fig:inference}~(a.2) and (b).
Intuitively, using clean histories ($t_{c}=0$) preserves maximum information.
However, this leads to the error accumulation issue, with generated artifacts propagated and amplified through subsequent samples.
We therefore examined the effect of conditioning timestep in the auto-regressive setting.
As shown in~\cref{fig:condition_noise}, substantial history details are preserved even with a relatively high timestep such as $t_c=500$, while at $t_c=900$, the overall structure remains though high-frequency details are lost.
Further analysis in~\cref{fig:ar_noise} demonstrates that low conditioning timesteps result in rapid quality degradation during sequential shot generation, while higher timesteps mitigate this problem at the cost to history fidelity.
Based on these findings, we employ $t_c=100 \sim 500$ in practice to optimize the balance between long-term generation quality and fidelity to historical context.

\begin{figure}
    \centering
    \includegraphics[width=\linewidth]{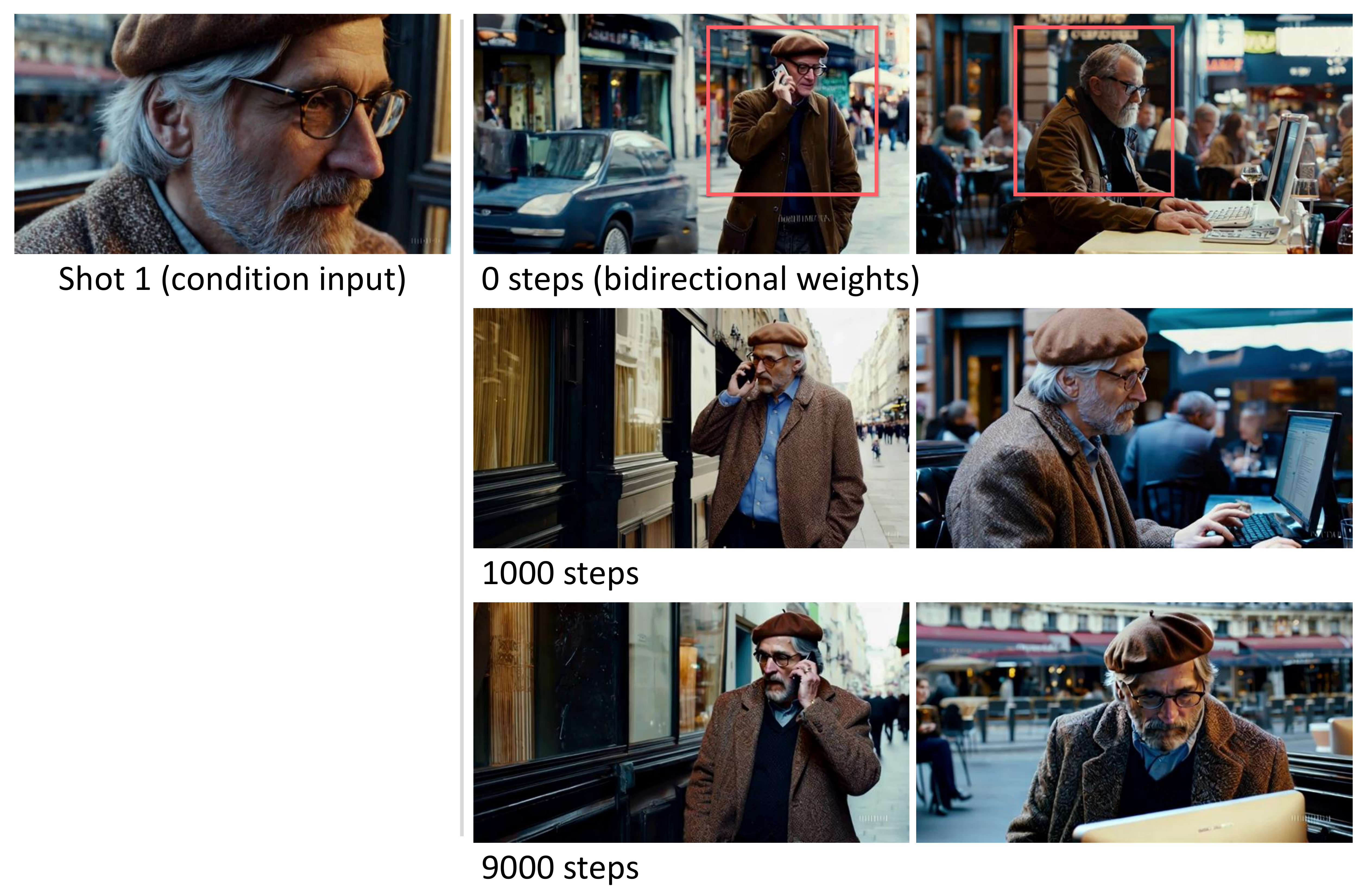}
    \caption{\textbf{Causal Adaptation.} After 1K updates from bidirectional weights, the causal architecture shows excellent consistency.}
    \label{fig:causal_adapt}
\end{figure}

\vspace{0.5em}
\noindent\textbf{Causal Fine-tuning Adaptation.}
We evaluate the efficiency of adapting a bidirectional model post-LCT to a context-causal architecture.
In~\cref{fig:causal_adapt}, we track performance evolution during adaptation by evaluating multiple checkpoints under identical conditions, with particular emphasis on identity consistency.
As illustrated, when initially loading bidirectional weights into the context-causal architecture (step 0), the model remains uncollaposed but generates frames with poor identity consistency.
After only 1K updates, identity consistency improves dramatically, achieving quality comparable to the final 9K-step checkpoint.
We attribute this efficiency to the bidirectional model's conditioning mechanism: the preceding clean history conditions rarely needs to attend to subsequent frames since they primarily contain their own information.
This behavior inherently resembles the unidirectional information flow enforced by the context-causal architecture, facilitating rapid adaptation between the two paradigms.

\subsection{Emerging Capabilities}
\label{sec:application}

In this section, we discuss several emerging model capabilities after Long Context Tuning~(LCT).

\vspace{0.5em}
\noindent\textbf{Conditional and Compositional Generation.}
Our model inherently supports diverse conditioning with images and videos.
As demonstrated in the second example of~\cref{fig:teaser}, we can generate narrative continuations by using an existing video as the initial shot condition.
Remarkably, despite no explicit training for this capability, our model enables compositional generation by accepting separate identity and environment images to synthesize coherent videos integrating these distinct elements, as illustrated in the final example of~\cref{fig:teaser}.
This ability emerges from the model's learned scene-level visual relations in the training corpus, where scenes frequently contain establishing environmental shots, character close-ups, and integrated shots depicting character-environment interactions.
Our findings offer a fresh perspective on appearance-conditioned video synthesis that eliminates task-specific data curation, providing a more generalizable approach to compositional generation.

\vspace{0.5em}
\noindent\textbf{Single-shot Extension.}
Benefit from training on single-shot segments~(\cref{sec:data_curation}), our model enables interactive shot extension without shotcut.
For this application, we remove the ``\texttt{[SHOT CUT]}" and craft bridging prompts that seamlessly connect existing content with desired future content.
As shown in the third example in~\cref{fig:teaser}, this approach can extend a single shot to minute-long durations, auto-regressively generating 10-second segments while maintaining visual consistency.

\vspace{0.5em}
\noindent\textbf{Interactive Generation.}
Our model enables interactive multi-shot development, as shown in the second example of~\cref{fig:teaser}.
This facilitates an iterative workflow where directors can progressively shape content shot-by-shot based on previously generated footages, eliminating the need for comprehensive upfront prompting and allowing for creative decision-making with immediate visual feedback.

\section{Conclusion}

We propose \textit{Long Context Tuning~(LCT)} to adapt single-shot video models for scene-level generation.
By expanding context window to the entire scene, implementing interleaved 3D position embeddings, and employing asynchronous training timesteps, we enable flexible scene synthesis without additional parameters.
Our approach also supports context-causal attention for efficient auto-regressive generation with KV-cache.
Evaluations show our method outperforms existing approaches in scene-level video generation while exhibiting emerging capabilities like compositional generation and shot extension.
We anticipate it will reveal new possibilities for video generation.

\vspace{0.5em}
\noindent\textbf{Discussion.}
As LCT allows users to create shots following their intents, we believe that in the future, leveraging the power of Multimodal Large Language Models (MLLMs) as planners for scene-level video generation would be a great alternative for further applications (\eg, story generation, even reasoning from generative perspectives).
Besides, current auto-regressive generation includes all preceding tokens as history conditions, which is effective yet introduces potential redundancy. Therefore, involving token dynamic routing into attention mechanisms may be a promising and practical direction for future work.

\clearpage

{
    \small
    \bibliographystyle{ieeenat_fullname}
    \bibliography{main}
}

\end{document}